# Spread spectrum compressed sensing MRI using chirp radio frequency pulses

Xiaobo Qu*, Ying Chen, Xiaoxing Zhuang, Zhiyu Yan, Di Guo, Zhong Chen*

*Abstract*—Compressed sensing has shown great potential in reducing data acquisition time in magnetic resonance imaging (MRI). Recently, a spread spectrum compressed sensing MRI method modulates an image with a quadratic phase. It performs better than the conventional compressed sensing MRI with variable density sampling, since the coherence between the sensing and sparsity bases are reduced. However, spread spectrum in that method is implemented via a shim coil which limits its modulation intensity and is not convenient to operate. In this letter, we propose to apply chirp (linear frequency-swept) radio frequency pulses to easily control the spread spectrum. To accelerate the image reconstruction, an alternating direction algorithm is modified by exploiting the complex orthogonality of the quadratic phase encoding. Reconstruction on the acquired data demonstrates that more image features are preserved using the proposed approach than those of conventional CS-MRI.

## I. INTRODUCTION

Magnetic resonance imaging (MRI) is widely applied in clinical diagnosis. Reducing data acquisition time is important for MRI since slow imaging speed leads to artifacts in images [1]. Assuming the image can be sparsely represented with a sparsity base, compressed sensing has shown promising results to accelerate the MRI [2]. Recently, a spread spectrum method outperforms conventional compressed sensing MRI (CS-MRI) with the variable density sampling, which is considered as the state-of-art sampling for a single image [3]. However, this spread spectrum achieved by quadratic phase modulation is implemented via a second order shim coil. Its implementation limits the modulation intensity and is not convenient to operate. In this letter, we introduce a chirp (linear frequency-swept) radio frequency (RF) pulses-based phase modulation [4, 5] into spread spectrum CS-MRI to simplify the measurement scheme. The modulation intensity is easily controlled by choosing a bandwidth of the chirp RF pulses. Observing that the complex orthogonality of the quadratic phase encoding matrix, a fast numerical algorithm is modified to reconstruct the image from undersampled data.

## II. SPREAD SPECTRUM COMPRESSED SENSING MRI

In this section, we give an overview of spread spectrum CS-MRI and explain the limitation of implementing spread spectrum with shim coil.

### A. Reduce the coherence using spread spectrum in CS-MRI

Let $\boldsymbol{\rho} \in \mathbb{C}^N$ denote an image, the data acquisition model for CS-MRI [2] is

$$\mathbf{s} = \mathbf{UF}\boldsymbol{\rho} + \boldsymbol{\eta},  \quad (1)$$

where $\mathbf{s} \in \mathbb{C}^M$ is the acquired data, $\boldsymbol{\eta} \in \mathbb{C}^M$ is the noise, $\mathbf{F} \in \mathbb{C}^{N \times N}$ is a Fourier basis matrix, and $\mathbf{U} \in \mathbb{R}^{M \times N}$ is an undersampling operator. A typical CS-MRI reconstruction attempts to solve

$$\hat{\boldsymbol{\alpha}} = \arg\min_{\boldsymbol{\alpha}} \|\boldsymbol{\alpha}\|_1 \ \ s.t. \ \|\mathbf{s} - \mathbf{UF}\boldsymbol{\Psi}\boldsymbol{\alpha}\|_2 \leq \varepsilon, \quad (2)$$

where $\boldsymbol{\Psi}$ is a dictionary and $\varepsilon$ stands for the noise level in the k-space data. The reconstructed image is $\hat{\boldsymbol{\rho}} = \boldsymbol{\Psi}\hat{\boldsymbol{\alpha}}$. According to the compressed sensing theory [6], the solution $\hat{\boldsymbol{\alpha}}$ is obtained if the number of measurements satisfies

$$M \geq C\mu^2(\mathbf{UF}, \boldsymbol{\Psi}) S \log N, \quad (3)$$

where $C$ is a constant, $\mu$ denotes the mutual coherence between encoding matrix $\mathbf{UF}$ and dictionary $\boldsymbol{\Psi}$, and $S$ is the number of nonzero entries in $\boldsymbol{\alpha}$. Reducing the coherence $\mu$ will reduce the image reconstruction error [3, 7].

Spread spectrum [3] was introduced into CS-MRI by modulating a quadratic phase $\boldsymbol{\Phi} \in \mathbb{C}^{N \times N}$ on image $\boldsymbol{\rho}$. $\boldsymbol{\Phi} \in \mathbb{C}^{N \times N}$ is a diagonal matrix with the diagonal entry $\phi(r_n) = e^{iar_n^2}$ where $r_n$ is the spatial location of $n^{\text{th}}$ voxel in the phase encoding direction and $a$ is a constant. The data acquisition model becomes

$$\mathbf{s} = \mathbf{UF}\boldsymbol{\Phi}\boldsymbol{\rho} + \boldsymbol{\eta}. \quad (4)$$

Compared with conventional CS-MRI [2], lower reconstruction error is achieved using spread spectrum [3]. The theoretical explanation is that the coherence $\mu(\mathbf{UF}\boldsymbol{\Phi}, \boldsymbol{\Psi})$ of spread spectrum is smaller than $\mu(\mathbf{UF}, \boldsymbol{\Psi})$ of conventional CS-MRI.

Manuscript received XX XX, 2013. This work was supported in part by the NNSF of China under Grants (61201045, 11174239, and 10974164). Asterisk indicates corresponding authors.

X.B. Qu, Y. Chen, X.X. Zhuang, Z.Y. Yan and Z. Chen are with Department of Electronic Science, Fujian Provincial Key Laboratory of Plasma and Magnetic Resonance, Xiamen University, Xiamen, 361005, China (e-mail: quxiaobo@xmu.edu.cn; pansychen@xmu.edu.cn; maikaz@qq.com; yanzhiyu.xmu@qq.com; chenz@xmu.edu.cn).

D. Guo is with School of Computer and Information Engineering, Xiamen University of Technology, Xiamen 361024, China (e-mail: guodi@xmut.edu.cn).

*B. Limitation of shim coil-based spread spectrum*

In the original spread spectrum MRI [3], the quadratic phase modulation on magnetizations was achieved by the second-order shim coils. Limitations of this scheme will be illustrated in this section.

Giving a gradient with linear spatial distribution $G(r) = G_0 + G_1 r$ acting over a temporal duration $T_0$, the phase variation satisfies

$$\Delta\phi(r) = \int_0^{T_0} \gamma \int_0^r G(r')dr'dt = \gamma G_0 T_0 r + \frac{\gamma}{2} G_1 T_0 r^2, \quad (5)$$

where $\gamma$ is the gyromagnetic ratio, and $r$ is the spatial location along the direction of the gradient. The quadratic phase modulation intensity is dominated by $G_1 T_0$.

However, the maximum magnitude of the shimming gradient is much lower than that of the imaging gradient, thus the spread spectrum effect provided by the original scheme will be limited since the modulation intensity is proportional to the curvature of the parabolic phase distribution [4, 5].

Using RF pulses is suggested as future work to simplify the measurement scheme of spread spectrum [3]. Recently, using chirp (linear frequency-swept) pulses shows the ability to more conveniently produce a much larger quadratic phase modulation [4, 5]. This method is immune to field inhomogeneity thus works better than single-shot echo planar imaging. It has not applied in CS-MRI as far as we know.

### III. PROPOSED METHOD

In this paper, we propose to apply chirp RF pulses in spread spectrum CS-MRI. In the following, the data acquisition model using chirp RF pulses is derived. Then, a fast image reconstruction algorithm is proposed to reconstruct images from undersampled data.

*A. Spread spectrum CS-MRI using chirp RF pulses*

The designed sequence is shown in Fig. 1. It is a variant of the conventional multi-scan spin-echo sequence [4, 5]. A main difference is that the common sinc pulse for excitation is replaced by a π/2 chirp pulse.

The chirp pulse has a linear frequency modulation as

$$\omega_{RF}(t_{enco}) = O_0 + R t_{enco}, \quad (6)$$

where $R$ and $O_0$ are the chirp rate and the initial frequency of the pulse, respectively. Bandwidth of this pulse is defined as $\Delta O = R T_{enco}$ [4, 5] for an excitation duration $T_{enco}$. For a field of view with length $L_Y$ in the phase encoding direction, relationship between $G_{enco}$ and $\Delta O$ should satisfy

$$\Delta O = \gamma G_{enco} L_Y, \quad (7)$$

since all the spins in the field of view are excited.

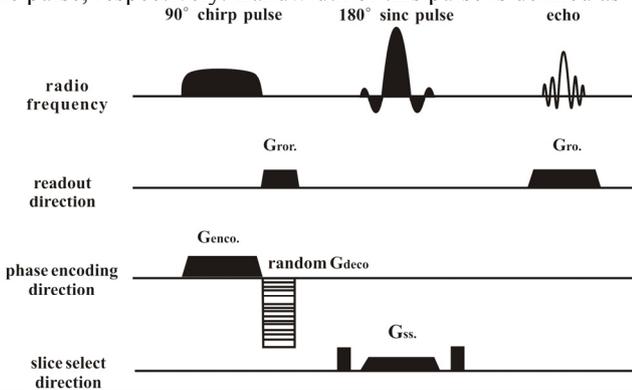

Fig.1. Pulse sequence for chirp RF pulses-based spread spectrum MRI.

In the excitation stage, a phase variation introduced by the chirp pulse at time instance $t_{enco}$ is

$$\varphi_{RF}(t_{enco}) = \int_0^{t_{enco}} (O_0 + R\tau)d\tau. \quad (8)$$

At the end of the excitation duration $T_{enco}$, a phase variation introduced by encoding gradient $G_{enco}$ is

$$\varphi_{G_{enco}}(t_{enco}) = \omega_{G_{enco}}(y)(T_{enco} - t_{enco}). \quad (9)$$

where $y$ is the spatial location in phase encoding direction.

At the end of encoding, the overall phase profile becomes

$$\varphi_{enco}(y) = \varphi_{RF}(t_{enco}) - \frac{\pi}{2} + \varphi_{G_{enco}}(t_{enco}). \quad (10)$$

This profile is further derived as

$$\varphi_{enco}(y) = -\frac{\gamma G_{enco} T_{enco}}{2L_Y} y^2 + \frac{1}{2}\gamma G_{enco} T_{enco} y - \frac{1}{8}\gamma G_{enco} T_{enco} L_Y - \frac{\pi}{2} \quad (11)$$

by substituting

$$\omega_{G_{enco}}(y) = \omega_{RF}(t_{enco}), \quad (12)$$

which means magnetization at the location satisfying Eq. (12) will flip from the Z-axis onto the X-Y plane at instance $t_{enco}$, and assuming

$$O_0 = -\frac{1}{2}\gamma G_{enco} L_Y, \quad (13)$$

which means the chirp pulse is symmetrically and incrementally swept [4, 5].

In the undersampling scheme, the phase variation generated by decoding gradient is

$$\varphi_{deco}(t) = -n\gamma \Delta g_{deco} t_{deco} y, \quad (14)$$

where $\Delta g_{deco}$ is the incremental magnitude of the decoding gradient between two successive scans, $t_{deco}$ is the duration of the gradient, and $n$ is the index of the scanning cycle chosen from $[1, N]$. $N$ is the required number of total scanning cycles for Nyquist sampling theorem using direct Fourier decoding, which satisfies

$$N = \frac{G_{enco} T_{enco}}{\Delta g_{deco} t_{deco}}. \quad (15)$$

The total phase distribution for the pulses is

$$\phi(y, m) = -\frac{\Delta O \cdot T_{enco}}{2L_Y^2} y^2 + \frac{m}{N} \cdot \frac{\Delta O \cdot T_{enco}}{L_Y} y - \frac{1}{8}\Delta O \cdot T_{enco} - \frac{\pi}{2}. \quad (16)$$

where $m = \frac{N}{2} - n$. Eq. (16) is derived by adding Eq. (11) and Eq. (14), and substituting Eq. (7).

Overall, the acquired signal for the sequence is

$$s(m) = \int_0^{L_Y} \rho(y) e^{i\phi(y,m)} dy = \int_0^{L_Y} \rho(y) e^{-ik_m y} \phi(y) dy, \quad (17)$$

where $k_m = -\frac{m}{N}\frac{\Delta O \cdot T_{enco}}{L_Y}$ and $\phi(y) = e^{-i\left[\frac{\Delta O \cdot T_{enco}}{2L_Y^2} y^2 + \frac{1}{8}\Delta O \cdot T_{enco} + \frac{\pi}{2}\right]}$.

By discretizing $y$ as $y_n = n\Delta y = n\frac{L_Y}{N}$ and scaling $s(m)$ with an appropriate factor, a discrete form of the data acquisition model in Eq. (17) is

$$\mathbf{s} = \mathbf{U}\mathbf{F}\tilde{\mathbf{\Phi}}\boldsymbol{\rho}, \quad (18)$$

where $\tilde{\mathbf{\Phi}} \in \mathbb{C}^{N \times N}$ is a diagonal matrix whose diagonal entry is

$$\tilde{\mathbf{\Phi}}_{n,n} = e^{-i\left[\frac{\Delta O \cdot T_{\text{enco}}}{2N^2}n^2 + \frac{1}{8}\Delta O \cdot T_{\text{enco}} + \frac{\pi}{2}\right]}. \quad (19)$$

$\tilde{\mathbf{\Phi}}$ stands for the quadratic phase modulated on an image. Multiplying $\tilde{\mathbf{\Phi}}$ with $\boldsymbol{\rho}$ corresponds to a convolution that generically spreads the spectrum of $\boldsymbol{\rho}$ [3].

Here, the modulation intensity is defined as

$$h = \frac{\Delta O \cdot T_{\text{enco}}}{N}, \quad (20)$$

since the phase modulation can be easily controlled by setting the value of $\Delta O \cdot T_{\text{enco}}$ for a fixed $N$. As shown in Fig. 2, the energy of k-space data is more widely spread out along the phase encoding direction for a higher $\Delta O$. The spread k-space data has the benefit to reduce the reconstruction error [3].

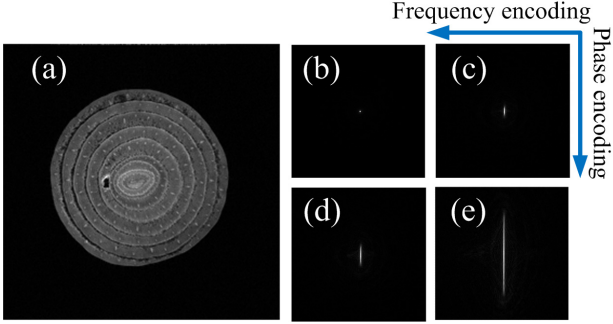

Fig. 2. Intensity of k-space for onion data at different chirp pulse bandwidth $\Delta O$. (a) magnitude image, (b)-(e) are the intensity of k-space when $\Delta O$ are 0kHz ($h$=0), 32kHz ($h$=0.125), 64kHz ($h$=0.25) and 256kHz ($h$=1), respectively. Note: $N$=256 and $T_{\text{enco}}$=4ms.

*B. Fast image reconstruction algorithm*

When the data $\mathbf{y}$ is undersampled, which means $M < N$, the Eq. (18) is underdetermined and reconstructing image is an ill-posed problem. For the randomly undersampled data, the compressed sensing MRI [2] is adopted to reconstruct the image

$$\hat{\boldsymbol{\rho}} = \arg\min_{\boldsymbol{\rho}} \left\{ \frac{\lambda}{2} \|\mathbf{s} - \mathbf{U}\mathbf{F}\tilde{\mathbf{\Phi}}\boldsymbol{\rho}\|_2^2 + \|\mathbf{\Psi}^H \boldsymbol{\rho}\|_1 \right\}, \quad (21)$$

where $\lambda$ trades data consistency with sparsity using transform $\mathbf{\Psi}^H$.

We adopt the alternating direction method (ADM) [8] to solve Eq. (21) because ADM enables fast computation in CS-MRI. The difference between spread spectrum MRI and conventional CS-MRI is the presence of $\tilde{\mathbf{\Phi}}$. We observe that $\tilde{\mathbf{\Phi}}$ satisfies complex orthogonality thus the ADM can make use of the special property of matrix to enable fast image reconstruction in spread spectrum MRI.

*Proof*: $\tilde{\mathbf{\Phi}}$ is a diagonal matrix with diagonal entry $\tilde{\mathbf{\Phi}}_{n,n} = e^{-i\left[\frac{\Delta O \cdot T_{\text{enco}}}{2N^2}n^2 + \frac{1}{8}\Delta O \cdot T_{\text{enco}} + \frac{\pi}{2}\right]}$. The Hermitian transpose of $\tilde{\mathbf{\Phi}}$ is also a diagonal matrix with diagonal entry $\tilde{\mathbf{\Phi}}^H_{n,n} = e^{i\left[\frac{\Delta O \cdot T_{\text{enco}}}{2N^2}n^2 + \frac{1}{8}\Delta O \cdot T_{\text{enco}} + \frac{\pi}{2}\right]}$. Therefore, $\tilde{\mathbf{\Phi}}\tilde{\mathbf{\Phi}}^H = \mathbf{I}$ since the diagonal entries of $\tilde{\mathbf{\Phi}}\tilde{\mathbf{\Phi}}^H$ satisfy $\tilde{\mathbf{\Phi}}_{n,n}\tilde{\mathbf{\Phi}}^H_{n,n} = 1$.

According to the ADM [8], an augmented Lagrangian form to solve Eq. (21) is

$$\min_{\boldsymbol{\alpha},\boldsymbol{\rho},\mathbf{v}} Q(\boldsymbol{\alpha},\boldsymbol{\rho},\mathbf{v}) = \frac{\lambda}{2}\|\mathbf{U}\mathbf{F}\tilde{\mathbf{\Phi}}\boldsymbol{\rho}-\mathbf{s}\|_2^2 + \|\boldsymbol{\alpha}\|_1 - \mathbf{v}^H(\boldsymbol{\alpha}-\mathbf{\Psi}^H\boldsymbol{\rho}) + \frac{\beta}{2}\|\boldsymbol{\alpha}-\mathbf{\Psi}^H\boldsymbol{\rho}\|_2^2. \quad (22)$$

where $\mathbf{v}$ and $\boldsymbol{\alpha}$ have the same dimension as $\mathbf{\Psi}^H\boldsymbol{\rho}$ and $\beta$ is a positive constant.

Eq. (22) is solved in an alternating fashion [8], which means three sub problems, estimating $\mathbf{v}$, $\boldsymbol{\alpha}$ and $\boldsymbol{\rho}$, will be solved alternatively until the solution converges. In the $g^{\text{th}}$ iteration, solving $\mathbf{v}$ and $\boldsymbol{\alpha}$ are the same as those were done in [8] and the derivations of solving them are omitted here. To demonstrate that complex orthogonality of $\tilde{\mathbf{\Phi}}$ enables fast computations, the following discussions focus on the derivation of solving $\boldsymbol{\rho}$ when $\boldsymbol{\alpha}$ and $\mathbf{v}$ are fixed.

When $\boldsymbol{\alpha}$ and $\mathbf{v}$ are given, the solution of

$$\min_{\boldsymbol{\rho}} Q_{\boldsymbol{\alpha},\mathbf{v}}(\boldsymbol{\rho}) = \frac{\lambda}{2}\|\mathbf{U}\mathbf{F}\tilde{\mathbf{\Phi}}\boldsymbol{\rho}-\mathbf{s}\|_2^2 - \mathbf{v}^H(\boldsymbol{\alpha}-\mathbf{\Psi}^H\boldsymbol{\rho}) + \frac{\beta}{2}\|\boldsymbol{\alpha}-\mathbf{\Psi}^H\boldsymbol{\rho}\|_2^2 \quad (23)$$

is the solution of a least square problem

$$\left(\beta\mathbf{\Psi}\mathbf{\Psi}^H + \tilde{\mathbf{\Phi}}^H\mathbf{F}^H\mathbf{U}^H\mathbf{U}\mathbf{F}\tilde{\mathbf{\Phi}}\right)\boldsymbol{\rho} = \beta\mathbf{\Psi}(\boldsymbol{\alpha}-\mathbf{v}) + \lambda\tilde{\mathbf{\Phi}}^H\mathbf{F}^H\mathbf{U}^H\mathbf{s}. \quad (24)$$

In the implementation, $\mathbf{\Psi}^H$ and $\mathbf{\Psi}$ denote the forward and inverse shift-invariant discrete wavelet transforms and they satisfy $\mathbf{\Psi}\mathbf{\Psi}^H = \mathbf{I}$. We adopt this wavelet transform since it can mitigate blocky artifacts introduced by orthogonal discrete wavelet in magnetic resonance image reconstruction [9]. The forward transform improves the shift-invariance by averaging over all possible shifts at computational cost $O(N \log N)$ for $N$-sample signals [10].

Since $\tilde{\mathbf{\Phi}}\tilde{\mathbf{\Phi}}^H = \mathbf{I}$ and $\mathbf{F}\mathbf{F}^H = \mathbf{I}$, multiplying $\mathbf{F}\tilde{\mathbf{\Phi}}$ on the left and right sides of Eq.(24) obtains

$$\left(\beta\mathbf{I} + \mathbf{U}^H\mathbf{U}\right)\mathbf{F}\tilde{\mathbf{\Phi}}\boldsymbol{\rho} = \beta\mathbf{F}\tilde{\mathbf{\Phi}}\mathbf{\Psi}(\boldsymbol{\alpha}-\mathbf{v}) + \lambda\mathbf{U}^H\mathbf{s}, \quad (25)$$

where $\mathbf{U}^H\mathbf{U}$ is a diagonal matrix with diagonal entries 0 or 1. The left side of Eq. (25) is invertible since $\beta > 0$, thus

$$\boldsymbol{\rho} = \tilde{\mathbf{\Phi}}^H\mathbf{F}^H\left(\beta\mathbf{I} + \mathbf{U}^H\mathbf{U}\right)^{-1}\left(\beta\mathbf{F}\tilde{\mathbf{\Phi}}\mathbf{\Psi}(\boldsymbol{\alpha}-\mathbf{v}) + \lambda\mathbf{U}^H\mathbf{s}\right). \quad (26)$$

Similar to the algorithm developed in [8], solving $\boldsymbol{\rho}$ only involves matrix-vector multiplications, fast Fourier transform and a fast sparsifying transform.

The pseudo code of the modified reconstruction algorithm is shown in Table I.

## IV. EXPERIMENTS AND RESULTS

In experiments, the full phase decoding number $N$=256 and the excitation duration $T_{\text{enco}}$=4ms. The field-of-view along Y-axis is $L_Y$=80mm. The repetition time is 1s and the echo time is 20ms. The acquisition rate is 50kHz. The acquired k-space data are shown in Figs. 3(a) and (b), which are undersampled in simulation according to the sampling pattern in Fig. 3(c). Simulations were run on a dual-core 2.2 GHz CPU laptop with 3GB of RAM. Computation time of image reconstruction is 8s.

TABLE I
IMAGE RECONSTRUCTION ALGORITHM FOR SPREAD SPECTRUM MRI

**Initialization:**
Input imaging parameters in Eq. (19) and the sampled data $\mathbf{s}$, set the $\beta = 2^8$ and regularization parameter $\lambda$, and initialize $\boldsymbol{\rho}^0 = \mathbf{F}^H \mathbf{U}^H \mathbf{s}$, $\mathbf{v}^0 = \mathbf{0}$, and iteration number $k = 1$.

**Main:**
  While $\|\boldsymbol{\rho}^{g+1} - \boldsymbol{\rho}^g\|_2 \geq 10^{-3} \|\boldsymbol{\rho}^g\|_2$, **Do**

  1. Update $\mathbf{v}$ according to $\mathbf{v}^{g+1} \leftarrow \mathbf{v}^g - \beta(\boldsymbol{\alpha}^g - \boldsymbol{\Psi}^H \boldsymbol{\rho}^g)$ ;
  2. Update $\boldsymbol{\alpha}$ according to $\boldsymbol{\alpha}^{g+1} \leftarrow H(\boldsymbol{\Psi}^H \boldsymbol{\rho}^g + \frac{1}{\beta} \mathbf{v}^{g+1}, \frac{1}{\beta})$ where $H(\mathbf{z}, \eta)$ is a soft thresholding on vector $\mathbf{z}$ with a threshold $\eta$;
  3. Update $\boldsymbol{\rho}$ according to Eq. (26);
  4. Update $g$ according to $g \leftarrow g + 1$;

**End Do**
**Output:** $\boldsymbol{\rho}^g = \boldsymbol{\Psi}\boldsymbol{\alpha}^g$

To evaluate the reconstruction, we use the relative $l_2$ norm error (RLNE) [9] defined as $e(\hat{\boldsymbol{\rho}}) = \|\boldsymbol{\rho} - \hat{\boldsymbol{\rho}}\|_2 / \|\boldsymbol{\rho}\|_2$ to measure the difference between the reconstructed image $\hat{\boldsymbol{\rho}}$ and the fully sampled image $\boldsymbol{\rho}$. The $\boldsymbol{\rho}$ was obtained by removing the phase modulation $\tilde{\boldsymbol{\Phi}}$ from $\tilde{\boldsymbol{\Phi}}\boldsymbol{\rho}$. A lower error implies that the reconstructed image is more consistent to the fully sampled image.

Less image features are lost in reconstruction (Fig. 3(f)) using the RF-based spread spectrum than those of using conventional imaging method [2] excluding spread spectrum (Fig. 3(i)). As shown in Fig. 4, when the modulation densities $h$ are $\{0.125, 0.25, 0.50\}$, which corresponds to $\Delta O = \{32, 64, 128\}$ kHz, lower reconstruction error is achieved using the proposed method. This observation is in accordance to that the spread spectrum MRI [3] obtains lower reconstruction error when the modulation densities lies in range of $[0.09, 0.30]$.

## V. CONCLUSION

A spread spectrum compressed sensing MRI using chirp radio frequency pulses is proposed. Connection between the designed pulse sequence and quadratic phase modulation is mathematically illustrated. The proposed method controls the intensity of spread spectrum more easily by setting chirp pulses bandwidth. Basing on the observation that quadratic phase modulation matrix satisfies complex orthogonality, an alternating direction method is modified to fast solve the reconstruction problem in 8s. Reconstruction on the acquired data demonstrates that more image features are preserved using the proposed approach than those of conventional CS-MRI.

ACKNOWLEDGMENT

The authors sincerely thank Dr. Congbo Cai, Mr. Jing Li, and Mr. Lin Chen for their kind help in this work. The authors aslo sincerely thank Drs. Gilles Puy, Junfeng Yang, Michael Lustig, and Richard Baraniuk for sharing their codes.

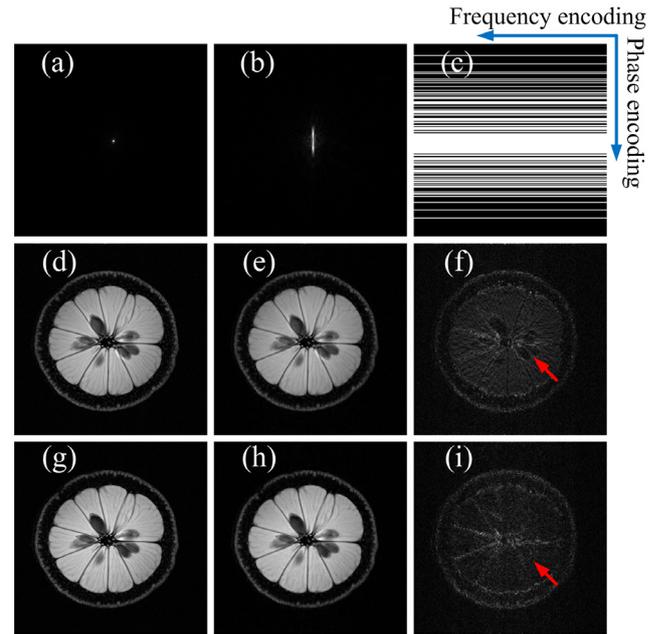

Fig. 3. Reconstructed images for lemon data. (a) and (b) are intensities of fully sampled k-space when $\Delta O$ is 0kHz ($h$=0) and 64kHz ($h$=0.25); (c) is the sampling pattern where the white pixel indicating the k-space data are sampled; (d) and (e) are reconstructed image of fully and 40% sampled data when $\Delta O$= 0kHz; (g) and (h) are reconstructed image of fully and 40% sampled data when $\Delta O$= 64kHz; (f) and (i) are the reconstruction error using undersampled data when $\Delta O$ is 0kHz and 64kHz. RLNEs for (e) and (i) are 0.082 and 0.069.

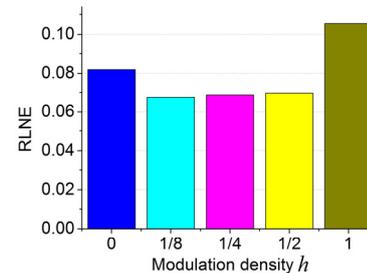

Fig. 4. Reconstruction errors at different modulation densities.